# Exploring the Frontiers of LLMs in Psychological Applications: A Comprehensive Review

**Luoma Ke[1], Song Tong[1,2*], Peng Cheng[3], Kaiping Peng[1,*]**

1. **Department of Psychological and Cognitive Sciences , Tsinghua University**
2. **Department of Psychology, Beijing Normal University**
3. **School of Social Science, Tsinghua University**

* Corresponding authors: tong.song.53w@kyoto-u.jp; pengkp@tsinghua,edu.cn

## Abstract

This review explores the frontiers of large language models (LLMs) in psychological applications. Psychology has undergone several theoretical changes, and the current use of artificial intelligence (AI) and machine learning, particularly LLMs, promises to open up new research directions. We aim to provide a detailed exploration of how LLMs are transforming psychological research. We discuss the impact of LLMs across various branches of psychology—including cognitive and behavioral, clinical and counseling, educational and developmental, and social and cultural psychology—highlighting their ability to model patterns, cognition, and behavior similar to those observed in humans. Furthermore, we explore the ability of such models to generate coherent, contextually relevant text, offering innovative tools for literature reviews, hypothesis generation, experimental designs, experimental subjects, and data analysis in psychology. We emphasize the importance of addressing technical and ethical challenges, including data privacy, the ethics of using LLMs in psychological research, and the need for a deeper understanding of these models' limitations. Researchers should use LLMs responsibly in psychological studies, adhering to ethical standards and considering the potential consequences of deploying these technologies in sensitive areas. Overall, this review provides a comprehensive overview of the current state of LLMs in psychology, exploring the potential benefits and challenges. We hope it can serve as a call to action for researchers to responsibly leverage LLMs' advantages while addressing the associated risks.

## 1. Introduction

Artificial intelligence (AI) has a history spanning nearly seven decades, beginning with the 1956 Dartmouth Conference. The field has recently been revolutionized with the advent of large language models (LLMs) such as ChatGPT, Google's Bard, and Meta's LLaMA. Among them, GPT-4, in particular, could signify a paradigm shift given its impressive capabilities (e.g., solving difficult tasks in math, coding, vision, medicine, law, and psychology) (Bubeck et al., 2023), exemplifying the concept of “AI for science” (Wang et al., 2023). LLMs mark a critical juncture in the evolution of machine learning and AI, propelled by their expansive size and sophisticated neural architectures that incorporate attention mechanisms (Vaswani et al., 2017). These models incorporate cognitive principles (Binz & Schulz, 2023a) and exhibit emergent properties comparable to those seen in complex physical systems (Wei et al., 2022). This has enhanced their ability to process and represent concepts and high-level semantics (J. Li et al., 2022) while also deepening our insights into human cognitive processes (Sejnowski, 2022). In psychological applications, these developments are reshaping interactions among data, language, and the environment (De Bot et al., 2007; Demszky et al., 2023), contributing significantly to various fields, including clinical (Thirunavukarasu et al., 2023), developmental (Frank, 2023; Hagendorff, 2023), and social psychology (Hardy et al., 2023; J. Zhang et al., 2023). Moreover, LLMs have had profound effects on psychological research methods, offering novel approaches and tools for exploration and analysis.

## 1.1. The LLM concept: From machine learning to capability emergence

Generative AI evolved from advances in pattern recognition capability. While convolutional neural networks (CNNs) excelled at recognizing objects and concepts, the next challenge was to use this recognition capability for a generation. For example, if a CNN can identify “age” in portraits, we can use that understanding to modify “age” in any portrait. This generative approach first succeeded in computer vision through models such as generative adversarial networks (Goodfellow et al., 2020) and deconvolution (Zeiler, 2014), which could create realistic images based on learned patterns. The same generative principles were then applied to language, leading to LLMs that could generate contextually relevant text. LLMs represent a particularly significant leap in the capabilities of generative AI. These models are designed to process natural language text and generate contextually relevant text. LLMs like GPT-4, LLaMA, Claude, and Gemini leverage the transformer architecture (Vaswani et al., 2017), which employs sophisticated neural networks and attention mechanisms to revolutionize natural language processing. Each model is optimized uniquely to enhance performance across diverse tasks. For instance, LLaMA focuses on efficient training processes (Touvron et al., 2023), Claude emphasizes safety and alignment (Li et al., 2024), and Gemini integrates advanced reasoning capabilities (Rane et al., 2024). LLaMA’s open-source nature allows local deployment and efficient training, making it suitable for psychological studies needing rapid iteration or customization, such as behavioral modeling (Binz and Schulz (2023a). Claude, designed for safety and alignment, is less commonly used in psychology research and more oriented toward knowledge-based tasks (Li et al., 2024). GPT-4, with its large-scale parameters and broad training data, supports a wide range of tasks, including cognitive simulations and clinical assessments. These differences guide model selection based on research needs like accessibility, task specificity, or data scale.

While these models highlight the versatility of LLMs, it is essential to distinguish between specific products designed for particular interactions, such as ChatGPT for conversational applications, and the broader capabilities of LLMs that extend beyond chat interfaces to include text generation, summarization, translation, and embedding extraction. This range of applications demonstrates that LLMs' capabilities are emergent, manifesting new abilities as the model size increases. Performance improvements on log-log scales sometimes experience "breaks" where unexpected capabilities emerge from complex interactions within the models (Wei et al., 2022).

At the heart of LLMs is the transformer architecture, a deep neural network with an attention mechanism that efficiently processes sequential data in parallel (Vaswani et al., 2017); this works in a manner somewhat similar to human brain functions. This architecture has revolutionized the field of natural language processing. The self-attention mechanism of the transformer architecture captures contextual relationships in textual data, allowing for more sophisticated language understanding. Notably, the "large" in LLM refers to the many parameters and massive amounts of training data used to fine-tune these models, typically billions of parameters and terabytes of text (Binz & Schulz, 2023b), in addition to "mastering the world" (Yildirim & Paul, 2023).

The process of large language modeling, from machine learning to the emergence of competence, can be divided into several key stages. (1) Pretraining: LLMs are pretrained on large amounts of textual data to learn intricate linguistic, syntactic, and textual structures, where the model learns to predict the next token through unsupervised learning, resulting in a base model that captures the statistical patterns of language (P. Liu et al., 2023). (2) Alignment: Supervised learning is used to create a foundation model that can better interact with users in the intended ways, which typically involves instruction tuning and reinforcement learning based on human feedback. After the foundation model is available, domain-specific fine-tuning can adapt the model for particular applications (Liu et al., 2022). This fine-tuning process ensures the model can generate contextually relevant responses and engage in meaningful conversations or tasks. Through these stages of development, LLMs demonstrate increasingly sophisticated text-generation capabilities, including response generation, content summarization, translation, and compositional text generation (Bubeck et al., 2023). The ability to effectively process and represent context is a critical factor underlying the observed emergence of advanced capabilities in these models. Finally, LLMs exhibit “observed capability emergence” when integrated into various applications and systems, in addition to performing tasks that require a deep understanding of language and context, thus often achieving human-like or superhuman performance in specific experimental tasks, such as analogical reasoning (Webb et al., 2023), creativity (Stevenson et al., 2022), and emotion recognition (Patel & Fan, 2023).

Therefore, LLMs can provide valuable insights into how such technologies can simulate or augment processes traditionally associated with human cognition. Specifically, LLMs maintain a balance between logical processing and the use of cognitive shortcuts (heuristics), and they adapt their reasoning strategies to optimize between accuracy and effort. This aligns with the principles of resource-rational human cognition, as discussed in dual-process theory (Mukherjee & Chang, 2024). For instance, LLMs generate and process natural language, demonstrating structural and functional parallels with certain aspects of human linguistic and cognitive mechanisms (Goertzel, 2023). These parallels allow for the exploration of AI applications in areas such as cognitive psychology (Sartori & Orrù, 2023), language acquisition (Jungherr, 2023), and even mental health (Lamichhane, 2023). Moreover, the study of LLMs contributes to our understanding of the human mind, offering a computational perspective on language processing, decision-making (Sha et al., 2023), and learning mechanisms (Hendel et al., 2023). The fusion of such disciplines could drive advancements in AI and provide a computational framework for investigating processes related to human cognition.

## 1.2. Psychology and AI

Psychology, as a science that explores the human mind and behavior, has undergone significant theoretical changes since the late nineteenth century, with psychoanalysis and behaviorism extending to cognitive psychology (Hothersall & Lovett, 2022). This history marks a shift in the focus of psychology research, reflecting the academic trend of moving from observing behavioral manifestations to exploring in-depth psychological connotations. Each of these phases has led to a deepening understanding of the psycho-cognitive processes of human beings.

Understanding human psycho-cognitive processes is therefore crucial for psychology. In clinical and counseling psychology, research on cognitive psychology supports diagnosing and treating psychological disorders. It deepens our understanding of the psychological mechanisms underlying emotions, stress, and human behavior. Psychotherapies such as cognitive-behavioral therapy (Hofmann et al., 2012) and psychodynamic therapy have become essential tools for promoting mental health and emotional regulation. In educational and developmental psychology, the development of cognitive psychology has fostered a deeper understanding of the roles of perceptual and affective factors in learning processes (Glaser, 1984), which has led to innovations in teaching methods and learning strategies. In social and cultural psychology, cognitive psychology research helps explain individuals' behaviors and mental processes in different social and cultural contexts, exploring how cultural differences affect cognitive patterns, values, and behavioral norms, especially in the context of globalization, interaction, and integration. In social psychology, meanwhile, cognitive psychology research on group behavior, social influence, prejudice, and discrimination holds great value for promoting social harmony and mutual understanding (Park & Judd, 2005).

AI is becoming an increasingly influential tool in psycho-cognitive research. Simon (1979) was among the first to recognize the potential of computational models to simulate aspects of human cognitive processes. Currently, LLMs can process and generate human-like texts and perform certain tasks in a manner similar to human cognition (Bubeck et al., 2023). LLMs also offer a unique computational perspective for the study of human cognition. For example, GPT-3 can solve vignette-based tasks similar to or better than human subjects and can perform rational decision-making based on descriptions, outperforming humans in the multiarmed bandit task (Binz & Schulz, 2023b). Furthermore, after extensive testing, GPT-3 is able to solve complex analogical problems at levels comparable to human performance, and analogical reasoning is an essential hallmark of human intelligence (Webb et al., 2023). Moreover, fine-tuning across multiple tasks can allow LLMs to predict human behavior in previously unseen tasks—that is, LLMs can be adapted to general-purpose cognitive models (Binz & Schulz, 2023a), potentially opening up new research directions that could transform cognitive psychology and behavioral science in general.

Newell (1990) offered a structured framework for analyzing human behavior, categorizing cognitive and behavioral processes into four distinct layers based on their time scales (Fig. 1a). At the biological level, the focus is on physiological and neural processes occurring at rapid time scales, ranging from milliseconds to one second. This level can include neural responses and sensory processing, which form the foundation of human cognition. The cognitive level pertains to mechanisms such as attention, perception, and short-term memory, which operate at intermediate time scales, typically between one second and one minute. These processes enable fundamental cognitive functions. At the rational level, the framework considers more complex cognitive activities such as problem-solving, planning, and decision-making. Such activities occur over longer time scales, spanning several minutes to a few hours, and involve sustained cognitive engagement. Finally, the social level considers behaviors shaped by social interaction and cultural influence, operating at the longest time scale, ranging from hours to days or longer. This level concerns the effects of social communication, group behavior, and cultural influences on cognition. It underscores the multifaceted nature of human behavior, highlighting the relationship between rapid physiological processes and the more prolonged, socially influenced aspects of human cognition. Figure 1 integrates this framework by mapping psychological domains (e.g., cognitive, social) onto these timescales, demonstrating LLMs' ability to simulate behaviors—from short-term processes like memory retrieval to long-term phenomena like cultural trends. Emergent properties (e.g., cognitive simulation) connect these domains to practical research tools (e.g., stimuli generation), with bidirectional influence refining both applications and properties.

Therefore, by analyzing LLM application across these four levels (Fig. 1a), it is possible to further explore their potential for modeling and studying human cognition and behavior (Fig. 1b), as well as their unique role in psycho-cognitive processes. Recent research has revealed significant advancements in LLMs' ability to perform complex human-like cognitive and social tasks (Grossmann et al., 2023; Marjieh et al., 2023; Orru et al., 2023; Pal et al., 2023; Stevenson et al., 2022; Webb et al., 2023). For instance, Grossmann et al. (2023) and Marjieh et al. (2023) demonstrated LLMs' proficiency in simulating human social interactions and perceptual processing, respectively. Orru et al. (2023) and (Webb et al., 2023) highlighted LLMs' capabilities in complex problem-solving and reasoning while Hagendorff et al. (2023) focused on decision-making processes. Stevenson et al. (2022) documented LLMs' potential for creativity, and Patel and Fan (2023) demonstrated their emotion-recognition abilities. Taken together, such findings highlight the expanding role of LLMs in representing and augmenting human cognitive and social functions, marking significant progress in AI research.

As general-purpose cognitive models (Binz & Schulz, 2023a), LLMs offer new perspectives and approaches for research in the fields of cognitive and behavioral psychology, clinical and counseling psychology, educational and developmental psychology, and social and cultural psychology, at different time scales of human behavior (Fig. 1a). LLMs can also be used as research aids (Fig. 1c) to help psychologists with everything from literature reviews (Aydın & Karaarslan, 2022; Qureshi et al., 2023), experimental subjects (Dillion et al., 2023; Hutson, 2023), and data analysis (Patel & Fan, 2023; Peters & Matz, 2023; Rathje et al., 2023) to promote scholarly communication: academic writing (Dergaa et al., 2023; Stokel-Walker, 2022) or peer review (Chiang & Lee, 2023; Van Dis et al., 2023). Thus, LLMs can potentially become research assistants for psychologists, helping them improve their research efficiency.

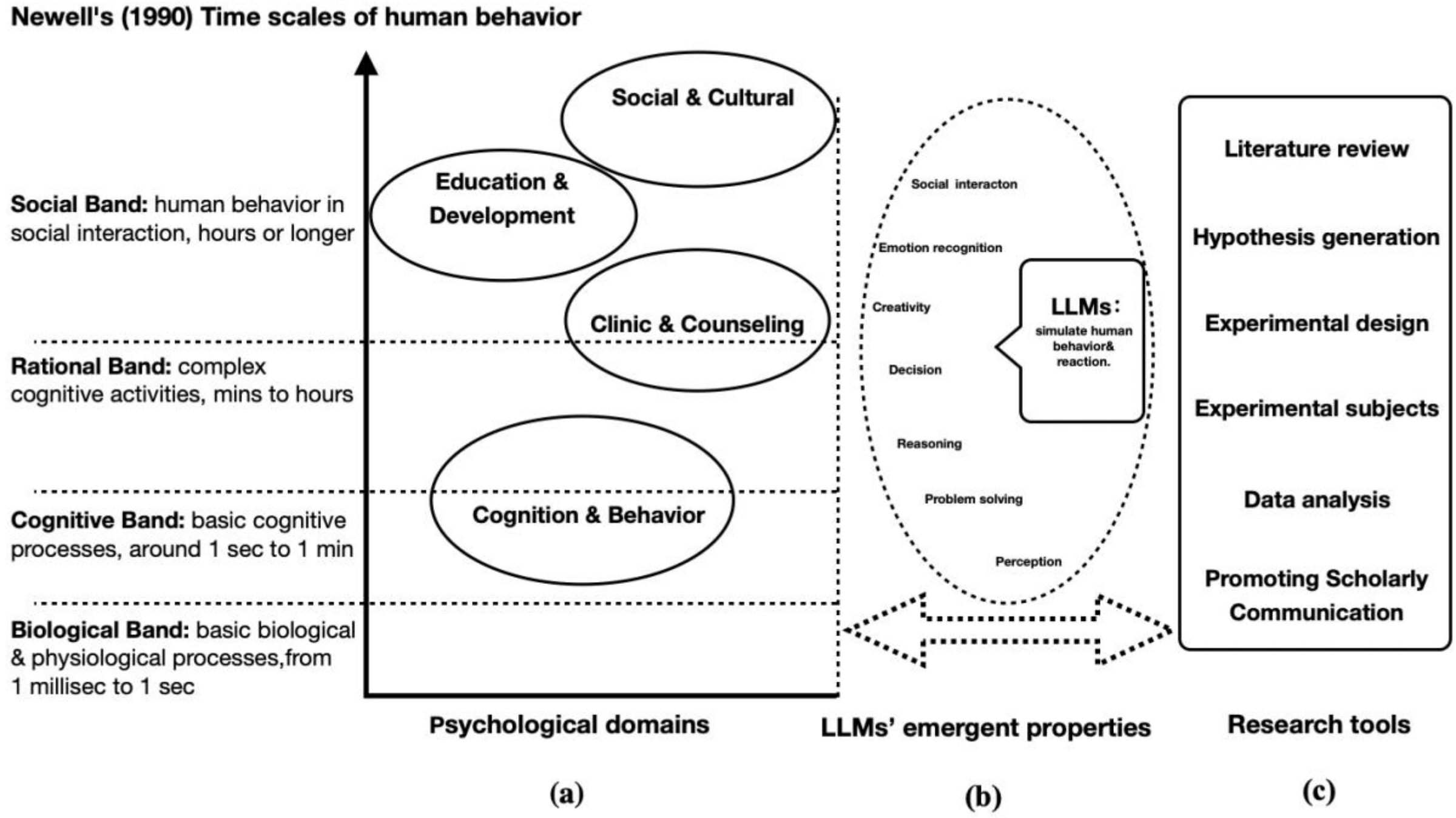

**Fig. 1 LLMs in Psychological Research Across Timescales.** (a) Domains (e.g., Cognitive & Behavioral, Social & Cultural) mapped to timescales of behavior; (b) Emergent properties (e.g., cognitive simulation) enabling domain-specific modeling; (c) LLMs as research tools (e.g., stimuli generation). Double-sided arrows indicate that emergent properties bridge domains and tools, supporting applications (e.g., memory retrieval) and refining properties through usage.

## 1.3. Objectives and significance of the present review

This review aims to provide a comprehensive analysis of the applications and effects of LLMs in psychological research. To ensure a systematic and rigorous review, we established specific inclusion and exclusion criteria. The review focuses on literature published between 2020 and 2024, sourced from relevant academic databases, including Google Scholar, arXiv, and Web of Science. Our initial keyword selection—"GPT-3", "ChatGPT", "GPT-4", "large language models", and "psychology"—was determined during the literature collection in October 2023, when GPT-based models were predominant in psychological research (e.g., Binz & Schulz, 2023b; Bubeck et al., 2023). At the time, open-source models like LLaMA (Touvron et al., 2023) and proprietary models like Claude had limited psychology-specific applications. To maintain a focused scope, we did not retrospectively expand search terms but included diverse LLMs via manual screening. To reflect recent developments, an updated search incorporated studies from 2024.

To bolster the integrity of our data extraction process, two interdisciplinary researchers (male, 33 and 41 years) specializing in information science and psychology conducted the encoding and screening. Our inclusion criteria required the selected studies to (1) explore the application or analysis of LLMs in psychological contexts; (2) be peer-reviewed journal articles or high-impact conference proceedings; and (3) present empirical data, theoretical discussions, or methodological advancements. We selectively included preprint articles if they addressed emerging trends or filled notable gaps in the literature. Articles without a psychological focus or those addressing non-LLM-based AI systems were excluded. The study selection involved screening 191 identified studies, analyzing 100 full-text articles, and ultimately including 47 studies categorized into various psychological subfields. Each of these studies met stringent inclusion criteria, ensuring they contributed meaningfully to our understanding of LLMs in psychological research.

In this review, we systematically examine the use of LLMs in various psychological domains by analyzing their application at different behavioral time scales. The rest of the paper is structured as follows. In section 2, we explore LLMs in cognitive and behavioral psychology. In section 3, the roles of LLMs in clinical and counseling psychology are discussed. Subsequently, educational and developmental psychology are addressed in section 4, followed by social and cultural psychology in section 5, outlining LLMs' contributions to each area. While psychological techniques are occasionally utilized to assess the capabilities of LLMs, this approach is employed to enhance understanding of their suitability and potential as instruments for psychological research. The primary focus of this review is how LLMs facilitate and advance psychological research across these domains. For a deeper understanding of the effect of LLMs on psychological research, an overview of LLMs' potential as tools for scientific research is given in section 6. In section 7, the challenges and future research directions with regard to applying LLMs to psychological contexts are provided. Finally, conclusions are presented in section 8, with a summary of LLMs' applications in psychology and recommendations for future work. Importantly, we propose strategies for integrating LLMs into psychological research and provide insights into interpreting such models from a psychological standpoint, contributing to their safety and interpretability.

## 2. LLMs in cognitive and behavioral psychology

Within the multilevel time scales of human behavior (Newell, 1990), cognitive and behavioral psychology has primarily focused on the study of cognitive processes at sub-hourly time scales, which encompass human engagement in perception, memory, thinking, decision-making, problem-solving, and conscious planning. Cognitive and behavioral psychology typically uses experimental methods to study these cognitive processes, controlling and observing behaviors and responses under specific conditions. The recent emergence of LLMs has reinvigorated the discussion on whether such models might exhibit patterns resembling human cognitive processes; if so, it may be possible to study the "cognitive processes" of LLMs, which could provide valuable insights into human cognitive phenomena and serve as a valuable addition to existing research methods in cognitive psychology. The foundational technology underlying large language models (LLMs) is the generative pre-trained transformer (GPT) architecture, which employs deep neural networks to process and generate human-like text. GPT models function through mechanisms, such as attention mechanisms and token prediction, enabling them to capture complex linguistic patterns and generate contextually coherent outputs. These foundational technologies have transformed natural language processing (NLP) by expanding the capacity for both comprehension and generation of text across diverse applications, from conversational agents to content creation (Brown et al., 2020; Vaswani et al., 2017). The incorporation of such architectures into psychological research has initiated discussions regarding their potential to simulate cognitive phenomena.

Binz and Schulz (2023a) found that fine-tuning multiple tasks enabled an LLM to predict human behavior in previously unseen tasks, suggesting that LLMs can be adapted to become generalist cognitive models. In another study, the same authors tested GPT-3 using tools from cognitive psychology and showed that it made better decisions than humans and outperformed them in the multiarmed bandit task (Binz & Schulz, 2023b). Other studies have shown that LLMs can display perceptual judgment (Marjieh et al., 2023), reasoning (Webb et al., 2023), and decision-making abilities (Hagendorff et al., 2023), as well as creativity (Stevenson et al., 2022) and problem-solving (Orru et al., 2023). One study found that an LLM had the mental ability of a seven-year-old child based on a false-belief task (which is considered the gold standard for testing theory of mind in humans) (Kosinski, 2024). Exploring the reasoning capabilities and decision-making processes of LLMs, Hagendorff et al. (2023) designed a series of semantic illusion and cognitive reflection tests designed to elicit intuitive but erroneous responses (these are conventionally used to study human reasoning and decision-making) and then ran the tests for LLMs. They conducted an analysis of model performance on a Cognitive Reflection Test (CRT) task and a semantic illusion task to elucidate their cognitive processes, drawing upon System 1 and System 2 thinking, as conceptualized by Daniel Kahneman in his seminal work Thinking, Fast, and Slow (Kahneman, 2011), which represent fundamental constructs for understanding human cognitive processes. System 1 refers to intuitive and automatic thinking, whereas System 2 involves rational, deliberate decision-making processes. This framework provides a theoretical basis for interpreting how LLMs simulate human-like cognitive behaviors during these tasks. They observe how these models show correct responses in these tasks and avoid pitfalls. The performance of the models in the CRT task were further evaluated by preventing them from chain-thinking to reason. The results showed that as model size and language capability increased, the LLMs increasingly exhibited human-like intuitive thinking (System 1) and the associated cognitive errors. Table 1 provides a summary of the applications of LLMs to cognitive and behavioral psychology.

**Table 1** Applications of large language models (LLMs) in cognitive and behavioral psychology study.

| Reference | Research Question | Research method | Key finding |
|---|---|---|---|
| ***Human-like Cognitive Abilities*** | | | |
| Binz and Schulz (2023b) | How does GPT-3 perform on cognitive psychology tasks, including decision-making, information search, and causal reasoning? | GPT-3 was tested using canonical cognitive psychology experiments and compared to human performance. | GPT-3 excels in decision-making and reinforcement learning but struggles with task perturbations, directed exploration, and causal reasoning. |
| Stevenson et al. (2022) | Can GPT-3 generate creative solutions comparable to humans in Guilford's Alternative Uses Test (AUT)? | GPT-3's responses to AUT were evaluated for originality, usefulness, surprise, and flexibility, using expert ratings and semantic distance analysis, and compared with human data. | Humans currently outperform GPT-3 in creativity, but GPT-3 shows potential to close the gap in future, raising questions about AI creativity and its evaluation. |
| Marjieh et al. (2023) | Can LLMs recover perceptual information from language, and how do they reflect cross-linguistic variations? | GPT-3, GPT-3.5, GPT-4 were tested on six psychophysical datasets and a multilingual color-naming task to compare their outputs with human perceptual data. | LLMs like GPT-4 align closely with human perceptual data, recover representations such as the color wheel, and reflect cross-linguistic perceptual variations, demonstrating their ability to extract perceptual information from language. |
| Loconte et al. (2023) | How do various LLMs perform on neuropsychological tests assessing prefrontal functions compared to human cognitive abilities? | GPT-3.5, GPT-4,were evaluated on tasks related to planning, semantic understanding, and Theory of Mind. | Findings indicate that GPT-4 generally meets normative human standards, whereas Claude2, and Llama2 show variable and often limited abilities, particularly in planning and Theory of Mind, underscoring the challenges in mimicking complex human cognitive functions. |
| Dhingra et al. (2023) | How does GPT-4 perform on cognitive psychology tasks compared to prior state-of-the-art models? | GPT-4 was evaluated on cognitive psychology datasets (CommonsenseQA, SuperGLUE, MATH, HANS) to analyze its integration of cognitive processes with contextual information. | GPT-4 demonstrates high accuracy on cognitive psychology tasks, surpassing prior models, and showcases significant potential to bridge human and machine reasoning. |
| Hagendorff (2024) | Can modern LLMs understand and apply deception strategies? | Experiments tested LLMs on inducing false beliefs, using chain-of-thought reasoning, and displaying Machiavellian behavior in simple and complex deception scenarios. | GPT-4 demonstrates advanced deceptive behavior, succeeding in 99.16% of simple and 71.46% of complex scenarios, highlighting the emergence of sophisticated deception abilities absent in earlier models. |
| ***Experimental Methodologies for Cognitive Research Using LLMs*** | | | |
| Binz and Schulz (2023a) | Can fine-tuned Mata's LLaMA accurately model human behavior? | LLaMA was fine-tuned on psychological experiment data and tested on decision-making tasks and unseen behaviors. | Fine-tuned LLaMA outperforms traditional cognitive models, accurately model individual behavior, and predict unseen human responses. |
| Dubey et al. (2024) | How can generative AI tools be utilized to streamline the creation of experimental stimuli in psychological research and influence public attitudes toward sustainable policies? | DALL-E 2 was employed to generate realistic visual stimuli of car-free urban environments, which were then presented to participants to measure attitudes toward sustainable policies. | By using DALL-E 2, the study demonstrated that generative AI tools can enhance the design process of experimental stimuli, offering greater control, diversity, and scalability, thereby effectively influencing participants' attitudes. |

**Note:** The AUT is a psychological test that measures creativity by asking participants to think of as many

uses as possible for a common object; DALL-E 2 is developed by OpenAI that generates detailed and realistic images from textual descriptions to explore AI's potential in creative fields.

Beyond theoretical evaluations, LLMs have demonstrated practical value in experimental psychology, particularly in stimulus generation and experimental design (Zhuang et al., 2023). Dubey et al. (2024), for instance, used DALL-E 2[1] to create realistic visual stimuli depicting car-free urban environments, which influenced participants' attitudes toward sustainable policies. Such tools streamline the stimulus design process by providing control, diversity, and scalability. Similarly, LLMs have been employed in hardware testing to generate tailored stimuli, outperforming traditional methods in specific scenarios (Z. Zhang et al., 2023). Charness et al. (2023) further demonstrated the use of LLMs for enhancing experimental workflows by refining task instructions, ensuring consistency, and monitoring participant engagement. By leveraging their flexibility and scalability, LLMs can provide novel methods for advancing experimental psychology. These applications facilitate the exploration of complex cognitive phenomena and the development of innovative research designs while also complementing traditional psychological research frameworks (Srinivasan et al., 2023). However, the interpretation of LLM outputs requires careful contextualization to avoid overstating their capabilities or equating them with human cognitive processes.

# 3. LLMs in clinical and counseling psychology

Clinical and counseling psychology focuses on assessing, diagnosing, treating, and preventing mental health problems. These processes often involve medium- to long-term periods. In the multilevel time scales of human behavior (Newell, 1990), clinical and counseling psychology involves assessing everyday behavioral acts (about a few hours to a day), habitual thinking (about a day to a few months), and psychological disorders (a few months to many years), among others (Fig. 1). The application of LLMs in clinical and counseling psychology can be broadly divided into two categories: psychological assessment and psychological intervention. Psychological assessment focuses on improving the ecological validity, scalability, and accuracy of measuring mental health states while psychological interventions consider how LLMs can be used for scalable and personalized mental health support, such as life coaching. According to related reports, there has been a public rush to use LLMs such as GPT for mental health screening and treatment (Demszky et al., 2023). LLMs are expected to be used in clinical psychology and counseling because they can parse human language and generate human-like responses, categorize text, and flexibly adapt conversational styles representing different theoretical orientations (Stade et al., 2023). This leads to the following question: How do LLMs work in psychotherapy, and can they replace human psychotherapists?

[1] Note: Although DALL-E 2 is not an LLM, we included this study due to its reliance on Transformer-based semantic understanding, a cornerstone of LLM research, and its demonstrated utility in generating controlled visual stimuli for psychological experiments.

An LLM is a basic generalized model with the ability to learn from small samples (Brown et al., 2020), which allows it to quickly become an "expert" in the clinical and counseling domain with only a small amount of data to learn from. For example, LLMs trained on clinical content can identify more specific factors of change that can help psychologists understand the process of clinical intervention, thus opening the black box of psychotherapy (Schueller & Morris, 2023). Regarding psychological assessment, studies have demonstrated that LLMs can effectively recognize emotions (Sharma et al., 2023) and respond appropriately. They can also perform complex mental health evaluations (Patel & Fan, 2023; Schaaff et al., 2023), such as suicide risk assessment and schizophrenia prognosis. For example, Elyoseph and Levkovich (2024) found that GPT-4, Google Bard, and Claude produced evaluations consistent with professional benchmarks in treated schizophrenia cases, though GPT-3.5 exhibited overly pessimistic predictions. Other research has shown that GPT-3.5 excels in clinical psychiatric cases, achieving top grades in diagnosis and management across 61% of cases, with only minor discrepancies in more complex scenarios (D'Souza et al., 2023).

In psychological interventions, LLMs have shown significant potential for delivering scalable and personalized mental health support (Blyler & Seligman, 2023a, 2023b). For example, Blyler and Seligman (2023a) demonstrated that GPT-4 can generate personalized therapeutic strategies by analyzing narrative identities. These strategies were found to be valid and reasonable, highlighting GPT-4's utility as a supportive tool in therapy and coaching. For peer-to-peer mental health support, Sharma et al. (2023) designed an AI system offering real-time empathic feedback, which improved overall empathy by 19.6% and significantly boosted self-efficacy among users struggling to provide support. Furthermore, J. M. Liu et al. (2023) evaluated ChatCounselor, a model trained on a domain-specific dataset of psychologist–client conversations, and found it outperformed open-source models, thereby demonstrating the importance of domain-specific training for improving counseling capabilities. Table 2 summarizes the applications of LLMs to clinical and counseling psychology.

The above research cases, which demonstrate LLMs' ability to provide clinicians with adequate mental health support (Schueller & Morris, 2023), hold promise for addressing insufficient capacity in the mental health care system and offering more individualized treatment services, even potentially fully automating psychotherapy in the future (Stade et al., 2023). It is essential to ensure that LLM is safe and privacy-protective in psychotherapy.

**Table 2** Applications of LLMs in clinical and counseling psychology studies.

| References | Research question | Research method | Key finding |
|---|---|---|---|
| ***Psychological Assessment Using LLMs*** | | | |
| Elyoseph and Levkovich (2023) | How effective and accurate are ChatGPT in assessing suicide risk? | The study evaluated ChatGPT's analysis of a text vignette depicting a hypothetical patient with varied psychological states, comparing its assessments to those of mental health professionals. | The study found that ChatGPT consistently underestimated suicide risk and mental resilience compared to mental health professionals, suggesting that reliance on ChatGPT for suicide risk assessment could lead to inaccurately low evaluations. |
| Elyoseph and Levkovich (2024) | How do LLMs compare to mental health professionals in assessing schizophrenia prognosis and treatment outcomes? | Vignettes were used to compare the assessments of four LLMs (GPT-3.5, GPT-4, Bard, Claude) against professional and public benchmarks. | GPT-4, Bard, and Claude aligned with professional views on treated cases, while GPT-3.5 was overly pessimistic. |
| D'Souza et al. (2023) | How effective is ChatGPT (3.5) in addressing clinical psychiatric cases and supporting mental health care? | ChatGPT was tested on 100 clinical psychiatric vignettes, and expert psychiatrists graded its responses across 10 categories. | ChatGPT excelled in management and diagnosis, earning top grades in 61% of cases, with no major errors but minor discrepancies in complex cases. |
| Sufyan et al. (2024) | How do the social intelligence (SI) levels of LLMs compare to human psychologists? | Social intelligence scores of ChatGPT (4), Bard, and Bing were compared with 180 counseling psychology students (bachelor's and PhD levels). | ChatGPT surpassed all psychologists in social intelligence, Bing outperformed most bachelor's and some PhDs, while Bard aligned with bachelor's students but fell behind PhDs. |
| ***Psychological Interventions with LLMs*** | | | |
| Blyler and Seligman (2023a) | Can GPT-4 generate personalized therapeutic strategies based on narrative identity? | GPT-4 analyzed five narrative identities to recommend tailored interventions. | GPT-4 effectively crafted personalized strategies, demonstrating its potential as a supportive tool for therapy and coaching. |
| Blyler and Seligman (2023b) | Can GPT-4 generate accurate and insightful personal narratives to support self-discovery in therapy and coaching? | GPT-4 processed 50 stream-of-consciousness thoughts from 26 participants to create personalized narratives, which participants evaluated for accuracy, surprise, and self-insight. | 96% of participants rated the narratives as accurate, and 73% reported gaining new self-insights, suggesting GPT-4's potential for enhancing self-discovery in therapeutic contexts. |
| Sharma et al. (2023) | Can AI enhance empathy in peer-to-peer mental health support? | Tested HAILEY which based on GPT-2, which offering real-time empathic feedback, in a trial with 300 peer supporters on TalkLife. | HAILEY improved empathy by 19.6% overall and 38.9% for those struggling with support, boosting self-efficacy without creating reliance. |
| J. M. Liu et al. (2023) | How to improve LLM in providing mental health support compared to other models? | ChatCounselor, trained on the Psych8k dataset of 260 psychologist-client conversations, was evaluated using the Counseling Bench with real-world counseling questions and psychological metrics. | ChatCounselor outperforms LLaMA, ChatGLM and approaches GPT-4's performance, highlighting the impact of domain-specific training on counseling capabilities. |

# 4. LLMs in educational and developmental psychology

Educational and developmental psychology is concerned with learning processes, knowledge accumulation, skill development, and changes in individual psychology in educational environments. Educational and developmental psychology is mainly positioned at the relatively medium- to long-term level (Newell, 1990), reflecting the ongoing learning and development that characterize the educational process. A national survey found that only 3 months after the public release of GPT, 40% of US teachers used it weekly for lesson planning, highlighting the growing impact of LLMs in education.

Table 3 summarizes the applications of LLMs in educational and developmental psychology, which can be broadly divided into two categories: developmental research with LLMs and using LLMs for education and learning applications. Here, developmental research seeks to determine whether LLMs can simulate human developmental processes (e.g., theory of mind and emotional reasoning) and how such capabilities might advance our understanding of human cognitive and emotional development. Kosinski (2024), for instance, tested different LLMs in 40 false-belief tasks and found that GPT-4 achieved 75% accuracy, comparable to the performance of a six-year-old child, while older models performed significantly worse. Vzorinab et al. (2024) used the Mayer–Salovey–Caruso Emotional Intelligence Test (MSCEIT) to evaluate GPT-4's emotional intelligence. While GPT-4 excelled at understanding and managing emotions, its reflective analysis resembled the early developmental stages of human emotional reasoning.

The study of using LLMs for education and learning applications focuses on leveraging LLMs to address challenges in education, such as providing personalized learning and improving learning motivation. LLMs learn from massive amounts of data taken from books and the Internet (Binz & Schulz, 2023b) and can be used as more knowledgeable learning aids (Stojanov, 2023), provide personalized learning experiences (Kasneci et al., 2023), and enhance the motivation to learn (Ali et al., 2023). Baillifard et al. (2024), for instance, found that an AI tutor powered by GPT-3 improved academic performance by up to 15 percentile points through personalized learning strategies. Stojanov (2023) used the following approach to explore GPT's potential as a learning tool: First, he set learning objectives and had "conversations" with GPT about its functionality for 4 hours. For the next 3 hours, he continued the discussion with GPT and watched some relevant videos on YouTube. He experienced positive feedback from his interactions with GPT and found it to be a motivating and relevant learning experience.

**Table 3** Applications of LLMs in educational and developmental psychology studies.

| References | Research question | Research method | Key finding |
|---|---|---|---|
| ***Developmental Research with LLMs*** | | | |
| Kosinski (2024) | Can LLMs demonstrate theory of mind (ToM) abilities? | Eleven different LLMs were tested on 40 false-belief tasks, requiring success in eight related scenarios per task. Performance was compared across model versions. | GPT-4 achieved 75% accuracy, comparable to 6-year-old children, while older models performed significantly worse. |
| Vzorinab et al. (2024) | How does GPT-4's emotional intelligence align with developmental patterns in human emotional reasoning? | GPT-4 was evaluated using the Mayer-Salovey-Caruso Emotional Intelligence Test (MSCEIT) through text-based prompts. | GPT-4 excels in understanding and managing emotions but shows limited reflective analysis, resembling early developmental stages in human emotional reasoning. |

| Trott et al. (2023) | How does exposure to language influence the development of theory of mind in humans and AI? | A linguistic False Belief Task was presented to humans and GPT-3 to assess belief attribution abilities. | GPT-3's partial success suggests that while language exposure contributes to belief reasoning, other developmental mechanisms unique to humans are crucial for fully developing theory of mind. |
|---|---|---|---|
| ***LLMs for Education and Learning Applications*** | | | |
| Stojanov (2023) | How effective is GPT as a learning aid in scaffolding understanding of a specific topic? | An autoethnographic study exploring the author's personal experience using ChatGPT(3.5) to learn about its technical aspects. | ChatGPT supports learning through motivating feedback but often provides superficial, inconsistent, and contradictory responses, risking overestimation of knowledge. |
| Jyothy et al. (2024) | What factors influence the adoption of LLMs like ChatGPT in learning, teaching, and research? | The Fogg Behavior Model (FBM) was applied to analyze the motivations, abilities, and perceptions of students, teachers, and researchers toward LLM use. | User motivation and ability drive LLM adoption, but limitations like teacher hesitance and technical challenges hinder broader integration. |
| Logacheva et al. (2024) | Can GPT-4 generate personalized programming exercises to enhance student engagement and learning? | GPT-4-generated exercises were evaluated in an introductory programming course by students and instructors for quality and engagement. | GPT-4 effectively produced high-quality, engaging exercises, offering personalized and scalable practice materials for programming education. |
| Machin et al. (2024) | Can GPT demonstrate psychological literacy comparable to subject matter experts (SMEs) in psychology research methods? | GPT rated 13 research scenarios, and its responses were statistically compared to SME evaluations. | GPT showed strong alignment with SME ratings ($r = .73–.80$), indicating its potential to match SME-level psychological literacy. |
| Ghafouri (2024) | Can a ChatGPT-based rapport-building protocol (CGRBP) enhance L2 (Second Language) grit in English learners? | A 16-week experimental study compared 30 EFL learners (15 experimental, 15 control) using pre-test post-test ANCOVA analysis. | CGRBP significantly improved L2 grit, demonstrating its potential to foster emotional support and learning motivation. |
| Ghafouri et al. (2024) | Can ChatGPT match expert psychological literacy in evaluating research methods? | The study compared responses from ChatGPT to 13 psychological research method scenarios against ratings by subject matter experts. | ChatGPT's responses correlated strongly with expert evaluations, suggesting its potential as an educational tool in psychology, though its usage should be approached with caution. |
| Baillifard et al. (2024) | Can AI tutors improve academic performance through personalized learning strategies? | A semester-long study with 51 psychology students using a GPT-3-powered AI tutor for personalized retrieval practice and progress modeling. | Active AI tutor use improved grades by up to 15 percentile points, with strong alignment between AI predictions and exam results. |

Note： Theory of mind (ToM) is the cognitive ability to attribute mental states to oneself and others； the Fogg Behavior Model (FBM) explains behavior as a product of motivation, ability, and prompts.

## 5. LLMs in social and cultural psychology

Social and cultural psychology explores how individuals interact with and are influenced by their social and cultural environments. It focused on interpersonal dynamics, group behavior, social cognition, and the long-term formation and transformation of attitudes and norms (Tajfel, 1982). Such phenomena occur at various time scales, from immediate social interactions to cultural changes evolving over several years (Newell, 1990). LLMs provide valuable tools for advancing social and cultural psychology. By analyzing textual datasets, simulating social interactions, and modeling human-like behaviors, LLMs can provide insights into the dynamics of social cognition, group processes, and cultural norms (Salah et al., 2023). Their scalability and ability to quantify patterns across time scales make them powerful instruments for examining human interactions in diverse contexts.

Research on LLMs in social and cultural psychology can be categorized into three main areas: cultural and cognitive understanding, social interactions and behavioral simulations, and practical applications. First, LLMs have many similarities with humans regarding social cognition. For example, research has found that LLMs reflect a variety of typical human cognitive biases in judgment and decision-making, such as the anchoring effect, the representativeness heuristic, and base-rate neglect (Talboy & Fuller, 2023). In addition, cultural psychology research has identified significant differences in the cognitive processes of Easterners and Westerners when processing information and making judgments (Nisbett et al., 2001); in this regard, LLM consistently favors holistic Eastern ways of thinking (Jin et al., 2023). Second, LLMs have been shown to characterize human groups in social interaction settings. It has been shown that LLMs can replicate the results of Milgram's electroshock experiments (Aher et al., 2023), show better gaming abilities in specific games (Akata et al., 2023), and exhibit different risk-taking and prosocial behaviors under different emotional states (Yukun et al., 2023).

Third, LLMs are increasingly used as proxies for human participants in psychological research. One study, for example, explored the potential of LLMs to serve as valid proxies for specific human subgroups in social science research; it found that LLMs contained information that went far beyond superficial similarity, reflecting the complex interplay between ideas, attitudes, and sociocultural contexts that characterizes human attitudes (Argyle et al., 2022). In addition, LLM has been tested for personality and values, obtaining results comparable to those for human samples, indicating their potential as psychological research tools (Miotto et al., 2022). Within this broader perspective, industrial and organizational psychology has increasingly employed LLMs, particularly in employee selection and workplace optimization, demonstrating their broader utility for understanding human behavior in structured environments. For example, LLMs have been shown to improve the accuracy and efficiency of recruitment systems in terms of assessing candidate fit and simulating workplace behaviors (Du et al., 2024). This approach can help mitigate biases and expand accessibility to a broader range of candidates. LLMs have also been integrated into systems such as PALR (personalization-aware LLMs for recommendation) to dynamically align individual capabilities with organizational needs. Such systems significantly reduce inefficiencies in hiring processes and enhance predictions about job performance by identifying nuanced compatibility factors in resumes and cover letters (Yang et al., 2023). Beyond individual applications, LLMs have contributed to understanding broader organizational cultures and transformational dynamics by providing insights into how group interactions and leadership styles influence workplace outcomes (Noy & Zhang, 2023). In the context of employee productivity, experiments using LLMs have revealed substantial benefits. For instance, professionals using ChatGPT in workplace writing tasks improved productivity by reducing task completion time by 40% and enhancing output quality by 18%, indicating its potential to effectively augment mid-level professional tasks (Noy & Zhang, 2023). Similarly, research on creativity has demonstrated LLMs' ability to help solve organizational problems requiring innovative thinking (Lee & Chung, 2024). Table 4 summarizes the applications of LLMs to social and cultural psychology.

LLMs have many applications in social and cultural psychology, allowing us to test theories and hypotheses about human behavior in social and cultural interaction settings. Zhao et al. (2024), for instance, examined whether AI chatbots can adjust their financial decisions and prosocial behaviors based on emotional cues, similar to humans. It was hypothesized that bots would take fewer risks when exposed to fear cues and more risks with joy cues. Emotional primes (fear, joy, or neutral) were applied, and investment decisions were analyzed. Additionally, prosocial responses, such as donating to a sick friend, were measured to assess how LLMs adapt behaviorally under emotional influences. These findings highlight LLMs' ability to model complex social dynamics and cultural influences. The next section broadens this perspective, exploring LLMs' potential as versatile research tools for psychologists.

**Table 4** Applications of LLMs in social and cultural psychology studies.

| References | Research question | Research method | Key finding |
|---|---|---|---|
| ***Cultural and Cognitive Understanding with LLMs*** | | | |
| Atari et al. (2023) | Do LLMs exhibit biases toward WEIRD (Western, Educated, Industrialized, Rich, Democratic) societies in psychological tasks? | LLMs' responses on psychological measures were compared to cross-cultural human data. | LLMs closely align with WEIRD cognitive patterns but show declining accuracy with non-WEIRD populations (r = -0.70), revealing a WEIRD bias. |
| Jin et al. (2023) | Does GPT exhibit cultural cognitive traits aligned with Eastern or Western thinking? | GPT was evaluated using cognitive and value judgment scales. | GPT leans towards Eastern holistic thinking in cognitive tasks but shows no cultural bias in value judgments, likely influenced by its training data and methods. |
| Schaaff et al. (2023) | How empathetic is GPT compared to humans? | GPT's empathy was evaluated through emotion recognition tasks, conversational analysis, and five empathy-related questionnaires. | GPT accurately identified emotions in 91.7% of cases, showed parallel emotions in 70.7%, and scored below average humans but above individuals with Asperger syndrome on empathy measures. |

| Patel and Fan (2023) | Can LLMs like Bard, GPT-3.5, and GPT-4 match human empathy and emotion identification? | Empathy and emotional understanding were assessed using TAS-20 (Toronto Alexithymia Scale-20) and EQ-60 (Emotional Quotient Inventory-60), comparing LLM responses to human benchmarks. | GPT-4 approached human-level emotional intelligence, outperforming Bard and GPT-3.5, which showed alexithymic tendencies. |
|---|---|---|---|
| X. Wang et al. (2023) | How do LLMs compare to humans in emotional intelligence? | A psychometric assessment focusing on Emotion Understanding was developed and applied to mainstream LLMs, benchmarking them against over 500 human participants. | GPT-4 scored higher than 89% of humans in emotional intelligence, with LLMs showing above-average emotional intelligence but using non-human mechanisms influenced by model design. |
| X. Li et al. (2022) | Are LLMs psychologically safe, and how can fine-tuning improve their safety? | LLMs were assessed using the Short Dark Triad (SD-3), Big Five Inventory (BFI), and well-being tests to evaluate personality traits and the impact of fine-tuning. | LLMs exhibit elevated dark traits but show improved well-being and psychological safety with targeted fine-tuning. |
| Miotto et al. (2022) | What are GPT-3's personality traits and values as assessed by validated psychological tools? | Administered validated personality and values measurement tools to GPT-3, including a model response memory to assess value alignment. | GPT-3 exhibits personality traits and values similar to human samples, providing initial evidence of psychological assessment in LLMs. |
| ***Social Interactions and Behavioral Simulations*** | | | |
| Zhao et al. (2024) | Can LLMs like GPT adapt responses to emotional primes in decision-making? | Tested GPT-4 and 3.5 with scenarios eliciting positive, negative, or neutral emotions. | GPT-4 showed distinct emotional response patterns, exceeding GPT-3.5, indicating advanced modulation but no true emotions. |
| Aher et al. (2023) | Can LLMs accurately simulate human behaviors, and what biases emerge in their simulations? | Introduced Turing Experiments to evaluate LLMs against findings from classic behavioral studies. | LLMs replicated most findings but showed "hyper-accuracy distortion," raising concerns for applications in education and the arts. |
| Abramski et al. (2023) | Do LLMs exhibit biases toward math and STEM, and how do these biases compare across models and with humans? | Using Behavioral Forma Mentis Networks, biases in GPT-3, GPT-3.5, GPT-4, and high school students were analyzed through a language generation task. | Newer LLMs (GPT-4) show reduced negative bias and richer semantic associations toward math and STEM compared to older models and humans, suggesting advancements in reducing stereotypes. |
| Almeida et al. (2024) | How do state-of-the-art LLMs reason about moral and legal issues, and how do their responses align with human judgments? | Eight experimental psychology studies were replicated using Google's Gemini Pro, Anthropic's Claude 2.1, GPT-4, and Meta's Llama 2 Chat 70b. Model responses were compared to human responses to assess alignment and systematic differences. | GPT-4 showed the best human alignment among LLMs but exaggerated effects and reduced variance, highlighting biases that limit their suitability as substitutes for human participants in psychological research. |
| ***Practical Applications with LLMs*** | | | |
| Noy and Zhang (2023) | How does generative AI affect employee productivity? | Workplace experiments with ChatGPT on writing tasks. | ChatGPT improved productivity (40% faster task completion, 18% better quality). |
| Lee and Chung (2024) | How does ChatGPT influence creativity? | Measured creativity using associative and divergent thinking tasks. | ChatGPT supports incremental creativity but is less effective for radical innovation. |
| Yang et al. (2023) | Can LLMs personalize job recommendations? | PALR(personalization-aware LLMs for recommendation) integrated interaction data with LLMs for dynamic recommendations. | PALR enhanced predictions of job performance and improved role matching. |
| Du et al. (2024) | How can LLMs improve job recommendations? | Used GANs with LLMs to refine low-quality resumes. | GAN-based systems predicted better job fit and reduced hiring inefficiencies. |
| Akata et al. (2023) | How do LLMs perform in social interaction tasks involving cooperation and coordination? | LLMs played repeated two-player games (e.g., Prisoner's Dilemma, Battle of the Sexes) with other LLMs and human-like strategies to analyze their behavior. | LLMs perform well in self-interest-driven games but struggle with coordination, with GPT-4 showing unforgiving behavior in the Prisoner's Dilemma and difficulty adopting simple coordination strategies. |
| Suri et al. (2024) | Does GPT exhibit heuristics and context-sensitive responses similar to those observed in human decision-making? | Four studies tested GPT's responses to prompts designed to assess cognitive biases (anchoring, representativeness, availability heuristic, framing effect, endowment effect) and compared them to human participant responses. | GPT demonstrated biases consistent with human heuristics across all studies, suggesting that language patterns alone may contribute to these effects, independent of human cognitive and affective processes. |
| Park et al. (2022) | How can designers predict and refine social behaviors in large-scale social computing systems before deployment? | Developed "social simulacra," an LLM-driven simulation that generates realistic community interactions based on design inputs (goals, rules, personas), allowing scenario testing and iterative design refinement. | Social simulacra accurately mimicked real community behavior, supported "what if?" scenario exploration, and helped designers improve system designs before large-scale deployment. |
| Sap et al. (2022) | Can LLMs demonstrate social intelligence and Theory of Mind (ToM)? | LLMs were evaluated using SocialIQa (social intents and reactions) and ToMi (mental states and realities), with results contextualized through pragmatics theories. | LLMs, including GPT-4, perform below human levels (55% on SocialIQa, 60% on ToMi), indicating that scaling alone does not yield ToM, highlighting the need for person-centric NLP approaches. |
| Argyle et al. (2022) | Can GPT-3 reliably emulate human subpopulations for social science research? | GPT-3 was conditioned on sociodemographic backstories from U.S. surveys, creating “silicon samples,” which were compared to human survey data. | GPT-3 exhibits nuanced, demographically aligned biases, indicating its potential as a tool for studying human behavior and societal dynamics. |
| P. S. Park et al. (2024) | Can GPT-3.5 simulate human participants and replicate social science study results? | Replicated 14 Many Labs 2 studies using GPT-3.5, analyzing response patterns and the “correct answer” effect through pre-registered and exploratory studies. | GPT-3.5 replicated 37.5% of study results but exhibited uniform responses (“correct answer” effect) and skewed conservative in moral foundation surveys, questioning its reliability and diversity as a human participant substitute. |

**Note:** WEIRD (Western, Educated, Industrialized, Rich, Democratic) refers to societies that represent a minority of the global population but are often overrepresented in psychological research.

# 6. LLMs as research tools in psychology

Sections 2-5 illustrate LLMs’ applications across cognitive, clinical, educational, and social psychology, revealing their potential to transform research practices. Together, these advancements speed up psychological research with new tools, encourage collaboration with fields like computer science and linguistics, and improve theoretical models through behavioral simulation—key ways LLMs advance psychology. Building on these foundations, this section explores LLMs as versatile research tools in psychology, supporting diverse tasks such as systematic reviews, literature review, and experimental design (Table 5). By reducing subjective bias and minimizing human variability in tasks like stimulus generation (Section 2), standardized assessments (Section 3), and data interpretation (Section 4), LLMs enhance objectivity and efficiency across these applications.

For instance, LLMs can automate systematic reviews and meta-analyses, revolutionizing evidence synthesis and provide actionable insights for psychologists, as grounded in cognitive and behavioral principles (e.g., in Sections 2 and 3). This capacity extends to enhancing psychologists’ workflows, building on productivity improvements noted in Section 5, through tools like literature review, hypothesis generation, experimental design, experimental subjects, and data analysis (Table 5).

**Table 5. LLMs as research tools in psychology study.**

| **Topic** | **Related study** |
|---|---|
| Literature review | LLMs can summarize the researched literature (Dis, Bollen, Zuidema, Rooij, & Bockting, 2023), complete literature review tasks (Qureshi et al., 2023), and create literature review articles (Aydın & Karaarslan, 2022), at the same time, there are LLM that has been specially trained to accomplish systematic literature reviews (Taylor et al., 2022)。 |
| Hypothesis generation | LLMs can generate hypotheses from scientific literature, make inferences based on scientific data, and then clarify their conclusions through interpretation (Zheng et al., 2023), and can quickly and automatically test these research hypotheses and learn from mistakes. |
| Experimental design | LLMs provide text-based material for experimental design, thereby optimizing the research process and reducing experimental complexity. By employing these models, researchers can easily create experimental stimuli, develop test items, and even simulate interactive sessions in controlled environments (Aher, Arriaga, & Kalai, 2022; Akata et al., 2023), providing a high degree of control and precision to the experimental process. |
| Experimental subjects | LLMs can simulate some human behaviors and responses, which provides an opportunity to test theories and hypotheses about human behavior (Grossmann et al., 2023), their use in place of human participation in experiments saves time and costs and can be applied to some experiments where human participation is not appropriate(Hutson, 2023), they can be combined with factors such as the specific research topic, the task, and the sample, and the use of LLM as an alternative to research participants where appropriate(Dillion et al., 2023). |
| Data analysis | LLMs can efficiently analyze massive amounts of textual data to gain insights into human behavior and emotions at an unprecedented scale (Patel & Fan, 2023), can analyze textual data in multiple languages, and accurately detect mental structures within it (Rathje et al., 2023), can draw mental profiles from social media data (Peters & Matz, 2023)。 |
| Scholarly Communication | LLMs can also help humans in writing(Dergaa et al., 2023; Stokel-Walker, 2022; Van Dis et al., 2023). LLMs were used in two natural language processing tasks and a human expert to assess the quality of the text, and the results of the assessment were consistent with those of the human expert (Chiang & Lee, 2023), LLMs offer the opportunity to get things done quickly, from Ph.D. students struggling to finish their dissertations, to peer reviewers submitting analyses under time pressure(Van Dis et al., 2023). |

## 6.1. Automated literature review and meta-analysis

Conducting a literature review meta-analysis is a complex, arduous process that requires significant time and expertise (Michelson & Reuter, 2019). Nature reported that researchers have used GPT as a research assistant to summarize literature (Dis et al. 2023). In one study, researchers used GPT to complete certain systematic literature review tasks (Qureshi et al., 2023). In another study, a literature review article was created using GPT with the application of digital twins in the health field; the results showed that knowledge compilation and representation were accelerated with the help of LLMs. However, their academic validity needs to be further verified (Aydın & Karaarslan, 2022). Researchers have also specifically trained LLMs to support the practical needs of scientific research (Taylor et al., 2022), including the ability to perform systematic literature reviews.

Recent studies have highlighted how LLMs can efficiently support meta-analysis. For instance, Luo et al. (2024) demonstrated that LLMs can screen literature, extract data, and generate statistical codes for meta-analyses, significantly reducing workload while maintaining recall rates comparable to manual curation. Similarly, Tong et al. (2024) used LLMs to extract causal pairs from 43,312 psychology articles, achieving an 86.98% success rate in pair extraction through adaptive prompting. As discussed in section 3, LLMs have shown strong capabilities in extracting causal relationships from large textual datasets, underscoring their potential to streamline evidence synthesis for systematic reviews and meta-analyses. Nevertheless, while LLMs excel in organizing qualitative data and identifying conceptual patterns, they face challenges in extracting the precise numerical data necessary for meta-analyses. For example, although LLM-based tools can retrieve and summarize outcome measures, manual validation remains essential to ensure accuracy, especially when processing complex figures or tables.

In summary, LLMs can speed up the process of literature review and meta-analysis. Researchers can use such models to systematically review and synthesize existing research, improving the efficiency of evidence-based psychology.

## 6.2. Hypothesis generation and experimental design

Hypothesis-driven research is at the core of scientific activity. LLMs can generate hypotheses from scientific literature, make inferences based on data, and then clarify conclusions through interpretation (Banker et al., 2024; Zheng et al., 2023). Although LLMs are capable of becoming "hypothesis machines," their logical and mathematical derivation capabilities still need improvement to eliminate factual errors, quickly test hypotheses, and learn from mistakes (Y. J. Park et al., 2024) . As innovative tools, LLMs have great potential for use in psychological experiments, given their ability to provide text-based material for experimental designs, thus optimizing the research process and reducing experimental complexity. Using such models researchers can easily create experimental stimuli, develop test items, and even simulate interactive sessions in controlled environments (Aher, Arriaga, & Kalai, 2022; Akata et al., 2023), providing a high degree of control and precision in the experimental process.

In conclusion, LLMs provide powerful, flexible tools for psychological research, from hypothesis generation to experimental design, which can help researchers achieve more precise, efficient research goals.

## 6.3. LLMs as subjects in psychological experiments

Although LLMs can simulate some human behaviors and responses—which provides an opportunity to test theories and hypotheses about human behavior (Grossmann et al., 2023)—there is still some controversy on whether LLMs can be used as a substitute for human subjects in psychological research. While recognizing that certain problems persist (e.g., biases and insufficiently trained data), some researchers have suggested that LLMs can be used as substitutes for human participants to save time and cost and can be applied to experiments that are not suitable for human participation (Hutson,2023). Others have proposed using LLMs as an alternative method of studying participants when appropriate, based on their performance in conjunction with factors such as specific research topics, tasks, and samples (Dillion et al., 2023). However, it is also believed that although LLMs can significantly affect scientific research, they are unlikely to replace human participants in any meaningful way (Harding et al., 2023). At the same time, some studies of LLMs as subjects have shown that LLMs perform similarly to humans (Orru et al., 2023; P. S. Park et al., 2024), which might indicate LLMs' potential to replace humans as subjects.

In conclusion, although LLMs can simulate human judgment, their simulation of human thinking remains limited, and their output should be validated and interpreted with caution when used as psychological subjects.

## 6.4. Tools for data analysis

Various forms of AI have long been used to analyze psychological data, such as flight data for pilot screening (Ke et al., 2023). Machine learning algorithms facilitate the processing of large datasets, identifying patterns and correlations that might otherwise be overlooked. However, LLMs take this capability to a new level; they can efficiently analyze massive amounts of textual data on an unprecedented scale to derive insights into human behavior and emotions (Patel & Fan, 2023). For psychological research, this means faster and more comprehensive data analysis, leading to more reliable, nuanced findings. LLMs can analyze textual data in multiple languages, accurately detect psychological structures within them (Rathje et al., 2023), and generate psychological profiles from social media data (Peters & Matz, 2023). LLMs have also demonstrated a degree of competence in the medical field; LLMs can, for example, predict the optimal neuroradiographic imaging modality for a given clinical presentation. Yet, LLMs cannot outperform experienced neuroradiologists, suggesting the need for continued improvement in the medical context (Nazario-Johnson et al., 2023). These findings demonstrate the great potential of LLMs for evaluating and analyzing data.

## 6.5. Promoting scholarly communication

Scholarly communication is a cornerstone of academic research, encompassing the processes of creating, evaluating, and disseminating knowledge. It includes writing research papers, conducting peer reviews, and ensuring that findings are communicated transparently and ethically. In psychology, this process is particularly complex owing to the field's diverse theoretical frameworks and methodological approaches, ranging from experimental to qualitative research. The discipline's focus on human behavior and its intersection with technology demands precise and ethical communication practices.

It has been argued that LLMs currently cannot completely replace human writing and instead can only answer questions and generate naturally fluent and informative content but with no real intelligence—i.e., text based on patterns of previously seen words (Stokel-Walker, 2022). In one study, students used GPT as an aid in their writing. The experimental group that used GPT was found to be similar to the control group in terms of writing quality, speed, and authenticity; the authors suggested that this could be because experienced researchers can better guide GPT to produce high-quality information. By contrast, students—who have less writing experience than researchers—found that GPT did not perform as effectively (Bašić et al., 2023). Another article discussed the prospects and potential threats of GPT in academic writing, emphasizing that using GPT in academic research should prioritize peer-reviewed scholarly sources. Yet, GPT's potential advantages for academic research, including the handling of large amounts of textual data and the automatic generation of abstracts and research questions (Dergaa et al., 2023), were highlighted. Furthermore, LLMs can potentially be used for peer review (Van Dis et al., 2023). The decisions/judgments of LLMs in a text-evaluation task were found to be consistent with those of human experts (Chiang & Lee, 2023).

In conclusion, LLMs such as GPT are potent tools for scholarly communication in psychology, capable of processing large amounts of textual data and automating tasks that were previously done manually. They can be used to scan academic papers and extract essential details, generate objective and unbiased abstracts, and create research questions in social psychology (Banker et al., 2023; Tong et al., 2024). However, researchers must exercise caution when using them as they can also introduce false or biased information into papers, leading to unintentional plagiarism and the misattribution of concepts (Van Dis et al., 2023).

# 7. Challenges and future directions

## 7.1. Challenges and limitations

LLMs have enormous potential to simulate complex cognitive processes, providing researchers with new tools to explore the mechanisms of human cognition and behavior for wide-ranging application in various fields, including clinical and counseling psychology, educational and developmental psychology, and social and cultural psychology. However, LLM output should not be mistaken for the presence of thought but instead viewed as complex pattern matching based on probabilistic modeling (Floridi & Chiriatti, 2020). Although LLMs show impressive performance, this differs from consciousness or genuine understanding. The interpretation of LLMs' capabilities must be based on an understanding of their limitations and the nature of their operations, which might differ fundamentally from human cognition. It is essential, then, to focus on the potential of LLMs in psychological research while also acknowledging the technical and ethical challenges that might arise.

First, despite the emergence of LLM competence (Wei et al., 2022), its internal working mechanism remains a black box from a cognitive and behavioral psychology perspective. For example, LLMs perform impressively on tasks requiring formal linguistic competence (including knowledge of the rules and patterns of a particular language) but fail many tests requiring functional competence (the set of cognitive abilities needed to understand and use language in the real world) (Mahowald et al., 2023). They excel in analogical and moral reasoning tasks but perform poorly on spatial reasoning tasks (Agrawal, 2023).

Second, while LLMs have accelerated the use of AI in clinical and counseling psychotherapy, privacy and ethical issues might arise (Graber-Stiehl, 2023). For example, gatekeepers, patients, and even mental health professionals who use GPT to assess suicide risk or improve decision-making might receive inaccurate assessments that underestimate risk (Elyoseph & Levkovich, 2023) or bias clinician decision-making, which can lead to healthcare inequities (Pal et al., 2023). In addition, LLMs in psychiatry research and practice have been associated with potential bias and privacy violations (Zhong et al., 2023).

Third, LLMs face application challenges in fields such as educational, developmental, and social and cultural psychology. It is evident that when applied in education, LLMs have the potential for output bias and misuse (Kasneci et al., 2023). One study found that texts generated by GPT were not always consistent or logical and sometimes even contradictory (Stojanov, 2023). In the field of social and cultural psychology, LLMs exhibit cognitive biases (Talboy & Fuller, 2023) and cultural biases (Atari et al., 2023) similar to those of humans, in addition to implicitly darker personality patterns (X. Li et al., 2022). Bender et al. (2021) suggested that training data for LLMs might reflect social biases that continue to be perpetuated in research settings.

Finally, LLMs have some limitations as aids to scientific research. With regard to writing, for example, LLMs currently cannot fully replace humans. Instead, they answer questions and generate naturally flowing, informative content lacking real intelligence (Stokel-Walker, 2022). Although macrolanguage models can simulate human judgment when used as experimental subjects, there are still limits to their "understanding" of human thought (Dillion et al., 2023). Van Dis et al. (2023) noted that LLMs might accelerate innovation, shorten publication times, and increase scientific diversity and equality. However, they might also reduce the quality and transparency of research and fundamentally alter scientists' autonomy as human researchers.

In summary, while LLMs offer extraordinary capabilities for psychological research, they also present challenges related to bias, ethical issues, data security, transparency, and technical expertise. Researchers should be fully aware of these challenges when using LLMs and adopt the following steps for ethical use:

First, disclose model details and methods transparently to ensure reproducibility. Second, verify outputs against literature or experts to address inaccuracies and misinformation. Third, use diverse training data to reduce cultural or gender biases. Fourth, in sensitive areas like mental health, limit use to assist—not replace—judgment and train users to interpret outputs critically. These steps, supported by recent studies (Abdurahman et al., 2024; Guo et al., 2024; Porsdam Mann et al., 2024), address ethical concerns in psychological research. Table 6 summarizes the challenges and limitations of LLMs in psychological applications.

**Table 6** Challenges and limitations of LLMs in psychological applications.

| Challenges | Author | Details |
|---|---|---|
| ***Cognitive and Behavioral Psychology*** | | |
| Lack of Real-World Understanding | Mitchell (2023) | LLMs lack real-world understanding, abstract reasoning, and intent comprehension. |
| Lack of Meta-knowledge | Stella et al. (2023) | LLMs fabricate information (hallucination) and lack curiosity/meta-knowledge. |
| Causal Reasoning and Creativity | Sartori and Orrù (2023) | Poor causal reasoning, dependence on biased training data, lack of creativity and imagination. |
| Multi-Step Reasoning Limitations | Goertzel (2023) | Poor multi-step reasoning, lack of autonomy, poor real-world understanding. |
| Common Sense Reasoning | Peng et al. (2023) | Forgetting knowledge in new tasks, poor common-sense reasoning, inconsistent problem-solving. |
| Model Behavior Challenges | Holtzman et al. (2023) | Lack of interpretability and formal behavioral descriptions makes systematic analysis difficult. |
| Psycholinguistic Features | Seals and Shalin (2023) | GPT and human-generated analogies differed in these stylistic dimensions, these lexical features, their choice of words for these features and these devices that help readers understand text. GPT may lack human cognitive and psycholinguistic features when generating analogies. |
| ***Clinic and Counseling Psychology*** | | |
| Technical Limitations& Patient Connection Issues | Stade et al. (2023) | Difficulty assessing suicide risk, substance abuse, safety issues, and interpreting nonverbal cues& Problems forming therapeutic relationships, interpreting nonverbal behaviors. |
| ***Education and Development Psychology*** | | |
| Integrity and Ethics Issues | Li et al. (2023) | Academic integrity concerns, misinformation, data privacy, and impact on critical thinking. |
| Bias and Over-Reliance& Multilingual Support Challenges | Kasneci et al. (2023) | Insufficient personalization, bias in teaching, over-reliance on models reduces creativity.& Limited support for diverse languages and equitable access. |
| ***Social and Culture Psychology*** | | |
| Liability and Privacy Issues | Fecher et al. (2023) | Liability issues: challenging traditional mechanisms of authorship and liability. Bias issues: affecting the objectivity and impartiality of science. Privacy and data protection issues: may be privacy issues with the training data of LLMs. Intellectual property issues: potential legal disputes. Environmental issues: generating large amounts of carbon emissions, which can have a negative impact on the environment. |
| Global Diversity Ignorance | Atari et al. (2023) | Ignoring global psychological diversity (e.g., tend to favor the psychological characteristics of WEIRD societies) and which can lead to prejudice and discrimination against people of other cultures and backgrounds. Differences in values and moral judgments and which can lead to problems of communication and understanding in multicultural societies. Self-identity and perceived social roles and which may lead to stereotypes and misconceptions about non-WEIRD populations). |
| Cultural and Ethical Tensions | P.S. Park et al. (2024) | Reduced innovation and development, bias and discrimination, culture clash and conflict, differences in values and morals and entrenchment of the status quo. |
| Social Context Limitations | Salah et al. (2023) | Limited understanding of social context: Although GPT performs well in syntax and general semantics, it still has limitations in capturing the nuances of social language. Ethical challenges: AI-generated fake content can lead to ethical issues including digital personhood, informed consent, potential manipulation, and the implications of using AI to simulate human interactions. |
| Bias and Misleading | Hayes | Potential biases: if the training data contain biases, LLMs may learn and replicate them. Data |

| Outputs | (2023) | privacy and consent issues: Text generated using LLMs may involve data privacy and consent issues. Output may be non-humanly understandable: although LLMs generate text that closely resembles human language, they do not truly understand the content and may generate absurd or misleading responses. |
| --- | --- | --- |
| Training Data Bias | Miotto et al. (2022) | Bias and discrimination: LLMs may be affected by biases in the training data, which can produce unfair results, such as reinforcing sexism in the translation of job advertisements. Responsibility and control: Due to the complexity of language models, it is difficult to determine who is responsible for the model's output, which can lead to attribution of problems and lack of controls. |
| Propagation of Harm | Bender et al. (2021) | Potential Harm: LLMs may lead to the propagation of harmful ideas such as stereotyping, discrimination, and extremism, and may lead to misinformation and bullying when generating text. Data bias and unfairness: leading to potential harm to marginalized communities. Automating bias: exacerbating existing biases and discrimination. Enhancement of authoritative viewpoints: LLMs may reinforce dominant viewpoints in the training data, further undermining marginalized people. |
| Alignment Challenges | Tamkin et al. (2021) | Alignment: In order to better align models with human values, algorithmic improvements are needed to increase factual accuracy and robustness against adversarial samples. In addition, appropriate values need to be made explicit for different usage scenarios. Societal Impact: Widespread use of LLMs may lead to problems such as information leakage and amplification of bias. |
| Misuse of LLMs | Brown et al. (2020) | Misuse of language modeling: GPT-3 may be used to generate fake news, spread extremist ideas, conduct cyber-attacks and other malicious uses. Fairness, bias, and representation: GPT-3 may carry bias against gender, race, and religion, among others, sparking related controversies. News generation: News generated by GPT-3 may be difficult to distinguish from real news, leading to confusing and misleading information. |
| ***Research Tools*** | | |
| Plagiarism and Copyright Issues | Sallam (2023) | Plagiarism: content generated by GPT may be considered plagiarized, violating academic norms. Copyright issues: Is the generated content owned by GPT or by the user? Transparency issues: The workings of GPT may not be transparent, making it difficult for users to understand the source of generated content. Liability issues: who is responsible for GPT when generating incorrect content? |
| Transparency Limitations | Gupta et al. (2023) | Transparency and Explanation: The working mechanism of generative AI models may be difficult to explain, which may lead users to doubt the credibility of the generated content. Legal and Ethical Issues: Generative AI models may involve intellectual property, privacy, and ethical issues, requiring attention to compliance with relevant laws and regulations during use. |
| Academic Integrity Concerns | Dergaa et al. (2023) | Integration of erroneous or biased information. Problems with citing original sources and authors. Impact on academic integrity and quality. Increased inequity and inequality: Difficulty in recognizing AI-generated content. Academic evaluation and recognition issues. Direct replacement for academic researchers: GPT is not a complete replacement for academic researchers as it has limitations in certain types of academic research. |
| Privacy and Bias Risks | Peters and Matz (2023) | User privacy: LLMs can infer psychological traits from a user's social media data, which may violate the user's privacy. Potential bias: LLMs may create potential bias in the inference process, which may lead to unfair treatment of specific groups (e.g., gender, age, etc.). Data security: if the inferential power of LLMs is used maliciously, it may lead to data leakage, with serious implications for users' mental health. |
| Misconduct and Limitations | Y. Liu et al. (2023) | Academic misconduct: GPT may be used for academic cheating, such as generating false papers or assignments. Challenges in the medical field: GPT has limitations in medical image analysis, which may lead to wrong diagnosis and jeopardize patients' health. |

## 7.2. Future directions and emergent trends

Currently, LLMs are used in different areas of psychology, including cognitive and behavioral, clinical and counseling, educational and developmental, and social and cultural psychology. As the capabilities of LLMs are further enhanced, their potential applications in psychology will continue to develop.

First, in the field of cognitive and behavioral psychology, with the emergence of multimodal LLMs (OpenAI, 2023), it is possible to combine visual and auditory information with textual data to better understand and model emotions, behaviors, and mental states for cognition. However, neuroimaging data can be used to inform the architectures and parameters of LLMs and integrate that information with traditional textual data to create more accurate and biologically sound models of human language and thought.

Second, in the field of clinical and counseling psychology, on the one hand, personal data, such as social media posts, medical records, or wearable device data, can be used to create tailored, personalized LLMs that provide more accurate and relevant insights into an individual's state of mind. At the same time, the strengths of human clinical and counseling expertise can be combined with the scalability and computational power of LLMs to create new diagnostic treatment and intervention tools. In addition, in educational and developmental psychology and social and cultural psychology, it is essential to build ethical LLMs and ensure they are designed and deployed in a way that respects privacy and uses data fairly and responsibly.

Ultimately, LLMs represent a systematic project whose future development cannot be achieved without the interdisciplinary collaboration of researchers in diverse fields such as psychology, computer science, and linguistics. For psychology researchers, accessible open-source LLM frameworks and tools might become an integral part of their future research efforts. Table 7 summarizes LLMs' future directions and emergent trends in psychological application.

**Table 7** Future directions and emergent trends of LLMs in psychological applications.

| **Author** | **Future directions and emergent trends** |
|---|---|
| ***Cognition and Behavior*** | |
| D'Oria (2023) | Delving into Human-Computer Interaction (HCI) to understand AI's ability to mimic human behavior. Exploring how AI language modeling can be applied in the human sciences to improve research efficiency and quality |
| Crockett and Messeri (2023) | Focus on the costs of adopting alternative human narratives in cognitive science research, such as masking the human labor behind them and the impact on human well-being. Concern about the impact of technological developments on scientific work and human understanding to ensure that cognitive scientists remain proactive in technological advances. |
| Binz and Schulz (2023b) | Explore ways to make LLMs more stable and robust in the face of descriptive tasks.<br>Investigate whether LLMs can learn to explore purposefully and how to better utilize causal knowledge in tasks. Analyze the performance of LLMs in different tasks and contexts to see if they can adapt like humans. Explore how LLMs develop and refine their cognitive abilities during natural interactions with humans. |
| Huang and Chang (2022) | Improve the reasoning ability of LLMs to encourage reasoning by optimizing training data, model architecture, and optimization goals. Develop more appropriate evaluation methods and benchmarks to measure the reasoning ability of LLMs to better reflect the true reasoning ability of the models. Investigate the potential of LLMs in different applications (e.g., problem solving, decision making and planning tasks). Explore other forms of reasoning (e.g., inductive and retrospective reasoning). |
| ***Clinic and Counseling*** | |
| Abd-Alrazaq et al. (2019) | Develop more chatbots for people with mental illness, especially for those with disorders such as schizophrenia, obsessive-compulsive disorder and bipolar disorder.<br>Implement more chatbots in developing countries to address the shortage of mental health professionals. Conduct more randomized controlled trials to evaluate the effectiveness of chatbots in mental health. |
| Stade et al. (2023) | Developing new therapeutic techniques and evidence-based practices (EBPs). Focus on evidence-based practices first: to create meaningful clinical impact in the short term, clinical LLM applications based on existing evidence-based psychotherapies and techniques will have the greatest chance of success. Involve interdisciplinary collaboration. Focuses on therapist and patient trust and usability. Criteria for designing effective clinical LLMs. |
| Demszky et al. (2023) | Development of high-quality cornerstone datasets: these datasets need to encompass populations and psychological constructs of interest and be associated with psychologically important outcomes (e.g., actual behaviors, mindfulness, health, and mental well-being). Focus on future research directions in consumer neuroscience and clinical neuroscience: research in these areas may involve the neural systems of marketing-related behaviors, decision neuroscience, neuroeconomics, and more. |
| ***Education and Development*** | |
| Hagendorff (2023) | Developmental psychology: examining how LLMs develop cognitively, socially, and emotionally over the lifespan and how these models can be optimized for specific tasks and situations. Learning psychology: studying how LLMs acquire and retain knowledge and skills, and how to optimize these models to improve learning. |
| ***Society and Culture*** | |
| Sap et al. (2022) | Explore more interactive and empirical training methods to help LLMs acquire true social intelligence and theoretical mental abilities. Investigate ways to combine static text with rich social intelligence and interaction data to improve social intelligence in LLMs. Investigate the theoretical-psychological abilities of LLMs in more naturalistic settings to reveal their performance in real-world scenarios. |
| Argyle et al. (2022) | Investigate the algorithmic fidelity of the GPT-3 model and how appropriate conditioning can allow the model to accurately simulate the response distributions of various human subgroups. Created "in silico samples" by conditioning on the socio-demographic backgrounds of real human participants in multiple large U.S. surveys. |
| Schaaff et al. (2023) | Developing more advanced models: to more accurately capture the emotional context of conversations and improve emotional understanding and expression. Measuring the emotional capabilities of bots: to investigate how to assess the emotional capabilities of chatbots in order to better understand how they behave when interacting with humans. Explore the use of GPT as a support tool: investigate how GPT can be used to support people more empathetically and improve human well-being. |
| Ziems et al. (2023) | Cross-cultural CSS research: future research should separately consider the utility of LLMs for cross-cultural CSS in order to better serve social science research in different cultural contexts. Future research could explore contrastive or causal explanations in LLMs. New paradigms for social science and AI collaboration. |
| ***Research Tools*** | |
| Van Dis et al. (2023) | Invest in truly open LLMs: develop and implement open-source AI technologies to increase transparency and democratic control. Embrace the advantages of AI: utilize AI to accelerate innovation and breakthroughs at all academic stages, while focusing on issues of ethics and human autonomy. Broaden the discussion: organize international forums to discuss the development and responsible use of LLMs in research, including issues of diversity and inequality. |
| Fecher et al. (2023) | Analyzing the risks and opportunities of LLMs for science systems. Examining how LLMs affect academic quality assurance mechanisms, academic misconduct, and scientific integrity. Exploring the impact of LLMs on academic reputation, evaluation systems, and knowledge dissemination. Examining how to balance the potential benefits from LLMs with adherence to scientific principles. |

# 8. Conclusion

With the rapid development of AI technologies, especially the continuous advancement of LLMs, machine learning has reached the point where it can recognize and generate human language. This development is not simply a technological breakthrough for the field of psychology, but it opens the door to a range of potential applications.

First, in the field of cognitive and behavioral psychology, LLMs are excelling in a variety of cognitive tasks. Although there are still limitations in causal cognition and planning, these models resurrect the principle of association, demonstrating the ability to associate at a distance and reason in complex ways. At the same time, the ability to adapt LLMs to cognitive models is a significant strength of psychological research, allowing for new explorations of human cognitive and behavioral processing mechanisms.

Second, in clinical and counseling psychology, LLMs can be used as preliminary diagnostic tools for mental health. While traditional mental health diagnosis relies on the experience of professionals and direct interaction with patients, LLMs can quickly identify potential mental health problems, such as depression and anxiety, by analyzing an individual's verbal expressions and textual content. Importantly, while such diagnoses cannot wholly replace professional psychological assessment, they can serve as an effective adjunct to help psychologists understand a patient's condition more quickly, or play a role in primary mental health interventions. Meanwhile, personalized psychological intervention is another critical application direction for LLMs. By combining information about an individual's health data and lifestyle habits, these models can provide tailored psychological advice and intervention programs. Such personalized approaches could be crucial for improving the effectiveness of psychological interventions.

Third, LLMs have the same potential for application in both educational and developmental psychology and social and cultural psychology. For example, LLMs provide interactive and personalized learning experiences or generate research tasks based on real-life applications that increase motivation and enhance learning. In addition, by analyzing large amounts of social media data, these models can help researchers track and analyze public sentiment changes to better understand psycho-social dynamics.

Finally, in psychological research, LLMs can drastically improve research efficiency. Researchers can use these models to quickly organize and analyze large amounts of literature, thus saving time. These models can also assist with experimental design, data analysis, and even promoting scholarly communication, making psychological research more efficient and precise.

In light of the above, LLMs have promising applications for psychology, such as research support, cognitive modeling, individualized intervention, and personalized learning. LLMs also have the potential to dramatically improve our understanding of human communication, thought processes, and behaviors, leading to the development of more comprehensive theories of mind and cognitive science. At the same time, it is important to be aware of the related risks and challenges and to ensure adherence to ethical standards, especially with regard to individual privacy and data security. It is also important to recognize that no matter how technologically advanced they are, LLMs can only partially replace the judgment and experience of human professionals. Therefore, such models should be viewed as an aid rather than an all-in-one solution.